\documentclass[journal]{IEEEtran}
\usepackage{amsmath,amssymb,amsfonts}
\usepackage{algorithmic}
\usepackage{graphicx}
\graphicspath{{figure/}}
\usepackage{textcomp}
\usepackage{comment}
\usepackage{multirow}
\usepackage{enumitem}
\usepackage{makecell}
\usepackage{rotating}
\usepackage{subcaption}
\usepackage{authblk}
\usepackage{color}

\newcommand{\RA}[1]{\textcolor{black}{#1}} 
\newcommand{\RB}[1]{\textcolor{black}{#1}}
\newcommand{\RC}[1]{\textcolor{black}{#1}}

\begin{document}

\title{Access Control of Semantic Segmentation Models Using Encrypted Feature Maps}

\author[]{Hiroki Ito}
\author[]{AprilPyone MaungMaung}
\author[]{Sayaka Shiota}
\author[]{Hitoshi Kiya}
\affil[]{Tokyo Metropolitan University, 6-6 Hino Tokyo, Japan}
\maketitle

\begin{abstract}
In this paper, we propose an access control method with a secret key for semantic segmentation models for the first time so that unauthorized users without a secret key cannot benefit from the performance of trained models.
The method enables us not only to provide a high segmentation performance to authorized users but to also degrade the performance for unauthorized users.
\RB{We first point out that, for the application of semantic segmentation, conventional access control methods which use encrypted images for classification tasks are not directly applicable due to performance degradation.}
Accordingly, in this paper, selected feature maps are encrypted with a secret key for training and testing models, instead of input images.
In an experiment, the protected models allowed authorized users to obtain almost the same performance as that of non-protected models but also with robustness against unauthorized access without a key.
\end{abstract}

\section{Introduction}
\label{sec:introduction}
Deep neural networks (DNNs) and convolutional neural networks (CNNs) have been deployed in many applications such as biometric authentication, automated \RA{driving}, and medical image analysis \cite{dl1, application0}.
However, training successful CNNs requires three ingredients: a huge amount of data, GPU-accelerated computing resources, and efficient algorithms, and it is not a trivial task.
In fact, collecting images and labeling them is also costly and will also consume a massive amount of resources.
Moreover, algorithms used in training a model may be patented or have restricted licenses.
Therefore, trained DNNs and CNNs have great business value.
Considering the expenses necessary for the expertise, money, and time taken to train a model, a model should be regarded as a kind of intellectual property (IP).

There are two aspects of IP protection for DNN models: ownership verification and access control \cite{kiya2022overview}.
The former focuses on identifying the ownership of the models, and the latter addresses protecting the functionality of the models from unauthorized access.
Ownership verification methods were inspired by digital watermarking \cite{watermark} and embed watermarks into models so that the embedded watermarks can be used to verify the ownership of the models in question \cite{embed0, embed1, embed3, embed7, embed5, embed4, embed6, embed9, embed2, embed8}.

Although the above watermarking methods can facilitate in identifying the ownership of models, in reality, a stolen model can be exploited in many different ways.
For example, an attacker can use a model for their own benefit without arousing suspicion, or a stolen model can be used for model inversion attacks \cite{inv-attack} and adversarial attacks \cite{adv1, adv2, adv3}.
Therefore, it is crucial to investigate mechanisms to protect DNN models from unauthorized access and misuse.
In this paper, we focus on protecting a model from misuse when it has been stolen (i.e., access control).

A method for protecting models against unauthorized access was inspired by adversarial examples and proposed to utilize secret perturbation to control the access of models \cite{access1}.
In addition, another study introduced a secret key to protect models \cite{access3}, and it was shown to outperform the other methods.
The secret key-based protection method uses a key-based transformation that was originally used by an adversarial defense in \cite{adv-def}, which was in turn inspired by perceptual image encryption methods \cite{pp1, pp2, pp3, pp5, pp6, etc1, etc2, etc4}.
This block-wise model protection method utilizes a secret key in such a way that a stolen model cannot be used to its full capacity without a correct secret key.
\RA{These existing methods provide a good access control performance, but they all focus on the access control of image classification models. In this paper, we point out that conventional access control methods with encrypted images for classification models are not directly applicable to segmentation models.}

Therefore, for the first time, in this paper, we propose a model protection method for semantic segmentation models by applying a key-based transformation to feature maps.
The method not only achieves a high classification accuracy (i.e., almost the same accuracy as in the non-protected case), but also increases the key space substantially.
\RA{Our contribution in this paper is to propose} an access control method with a secret key for semantic segmentation models for the first time, which enables us not only to maintain a high segmentation accuracy but also to increase the key space.
To evaluate the proposed method, we conduct relevant attacks.
In experiments, the proposed model-protection method is confirmed to outperform previous such methods.

\section{Related Work}
\label{sec:related}
\RA{There are two approaches to protecting trained models: ownership verification and access control. The former focuses on identifying the ownership of trained models. The latter addresses protecting the functionality of trained models. The former aims for only ownership verification. Therefore, stolen models can be directly used by unauthorized users, so we focus on access control to protect trained models from unauthorized access even if the models are stolen.}

The first access control method, which was inspired by adversarial examples \cite{adv1, adv2, adv3}, was proposed for image classification models in \cite{access1}. 
In this method, authorized users add a secret perturbation generated by an anti-piracy transform module to input images, and the processed input images are fed to a protected model.
Therefore, this method needs additional resources to train the module.
In addition, the method focuses on protecting image classification models.

The second method is to extend the passport-based ownership verification method \cite{embed5} as an access control method.
However, the passport in \cite{embed5} is a set of extracted features of a secret image/images or equivalent random patterns from a pre-trained model.
In addition, a network has to be modified with additional passport layers to use passports.
Therefore, there are significant overhead costs in both the training and inference phases.
Moreover, the effectiveness of the passport-based method has never been confirmed under the use of semantic segmentation models.

\begin{figure}
    \centering
    \begin{subfigure}{0.2\columnwidth}
        \centering
        \includegraphics[width=\columnwidth]{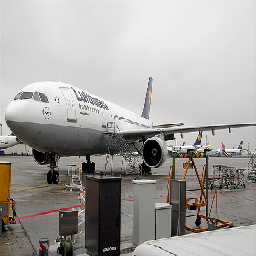}
        \caption{Original}
        \label{fig:original_sample}
    \end{subfigure}
    \begin{subfigure}{0.2\columnwidth}
        \centering
        \includegraphics[width=\columnwidth]{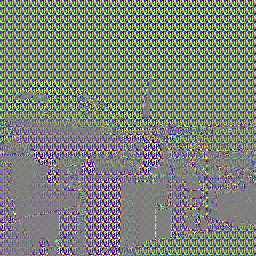}
        \caption{NP}
        \label{fig:np_sample}
    \end{subfigure}
    \begin{subfigure}{0.2\columnwidth}
        \centering
        \includegraphics[width=\columnwidth]{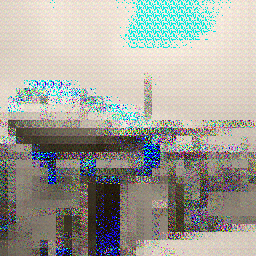}
        \caption{SHF}
        \label{fig:shf_sample}
    \end{subfigure}
    \begin{subfigure}{0.2\columnwidth}
        \centering
        \includegraphics[width=\columnwidth]{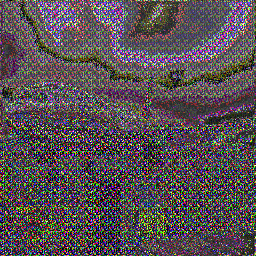}
        \caption{FFX}
        \label{fig:ffx_sample}
    \end{subfigure}
    \caption{Images transformed by block-wise transformations}
    \label{fig:block_sample}
\end{figure}

\begin{figure}
    \centering
    \includegraphics[width=0.8\columnwidth]{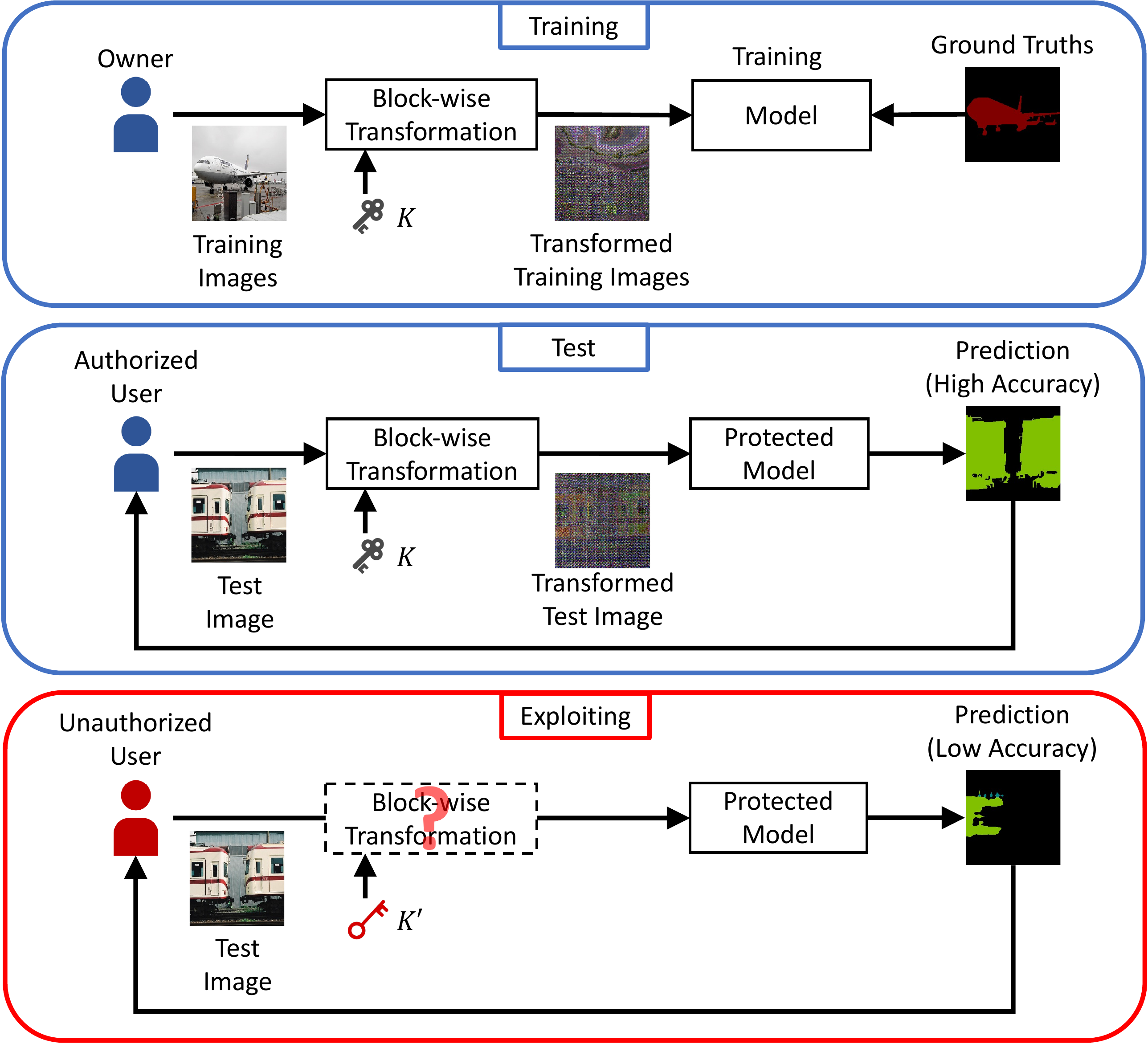}
    \caption{Access control framework with encrypted input images}
    \label{fig:block_overview}
\end{figure}

The third is a block-wise image transformation method with a secret key \cite{access3}, which is inspired by learnable image encryption \cite{adv-def,pp1,pp2,pp3,pp5,pp6}.
In this method, input images are encrypted with a key, for which three types of encryption methods: negative/positive transformation (NP), pixel shuffling (SHF), and format-preserving Feistel-based encryption (FFX), were proposed as illustrated in Fig.~\ref{fig:block_sample}.
Figure~\ref{fig:block_overview} shows the framework of the block-wise method.
In the framework, an owner transforms all training images with secret key $K$, and a model is trained to protect the model by using the transformed images and corresponding ground truths.
An authorized user with key $K$ transforms a test image with $K$ and feeds it to the protected model to get a prediction result with high accuracy.
In contrast, an unauthorized user without key $K$ cannot obtain a prediction result with high accuracy, even if the unauthorized user knows the framework and the encryption algorithm.
In addition, the method with key $K$ does not need any network modification or incur significant overhead costs.
However, the use of the block-wise transformation is limited to image classification tasks.

Accordingly, in this paper, we propose a novel access control method for semantic segmentation tasks for the first time.
The proposed method does not need any network modification or incur significant overhead costs as well.

\section{Access Control with Encrypted Feature Maps}
\label{sec:protect}
Access control of semantic segmentation models with encrypted feature maps is proposed here.

\subsection{Overview}
\begin{figure}
    \centering
    \includegraphics[width=0.8\columnwidth]{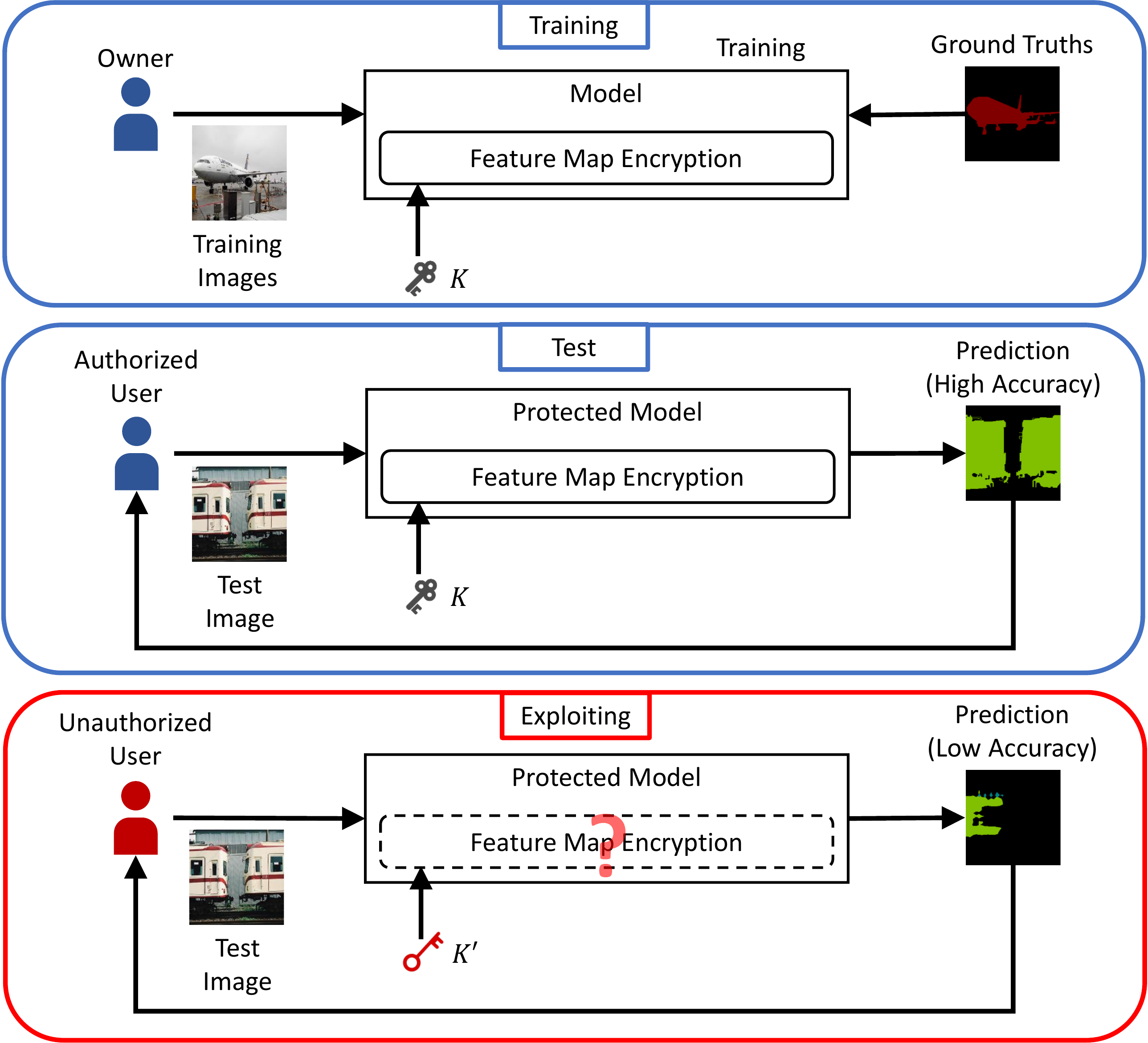}
    \caption{Access control framework with encrypted feature maps}
    \label{fig:feature_overview}
\end{figure}

Protected models for access control should satisfy the following requirements.
The protected models should provide prediction results with a high accuracy to authorized users but not provide such high-accuracy results to unauthorized users. 
To meet these requirements, encrypted feature maps are used as shown in Fig.~\ref{fig:feature_overview}.

In the framework with encrypted feature maps, an owner trains a model by using plain training images and corresponding ground truths, where selected feature maps in the network are encrypted by using a secret key $K$ at each training iteration in accordance with the proposed method.
For testing, an authorized user with key $K$ feeds a test image to the trained model to obtain a prediction result with high accuracy.
In contrast, when an unauthorized user without key $K$ inputs a test image to the trained model without any key or with an estimated key $K'$, the unauthorized user cannot benefit from the performance of the trained model.

\subsection{Feature Map}
\begin{figure*}
    \centering
    \begin{subfigure}{15cm}
        \centering
        \includegraphics[height=4cm]{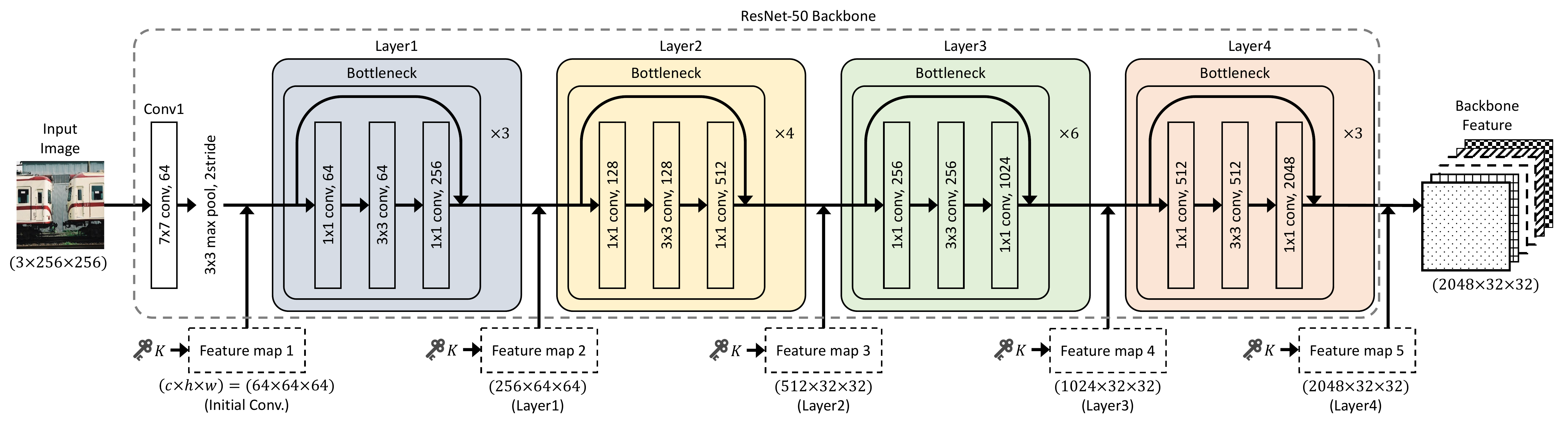}
        \caption{Backbone for FCN and DeepLabv3 (ResNet-50 architecture)}
        \label{fig:backbone}
    \end{subfigure} \\
    \begin{subfigure}{5.25cm}
        \centering
        \includegraphics[height=4cm]{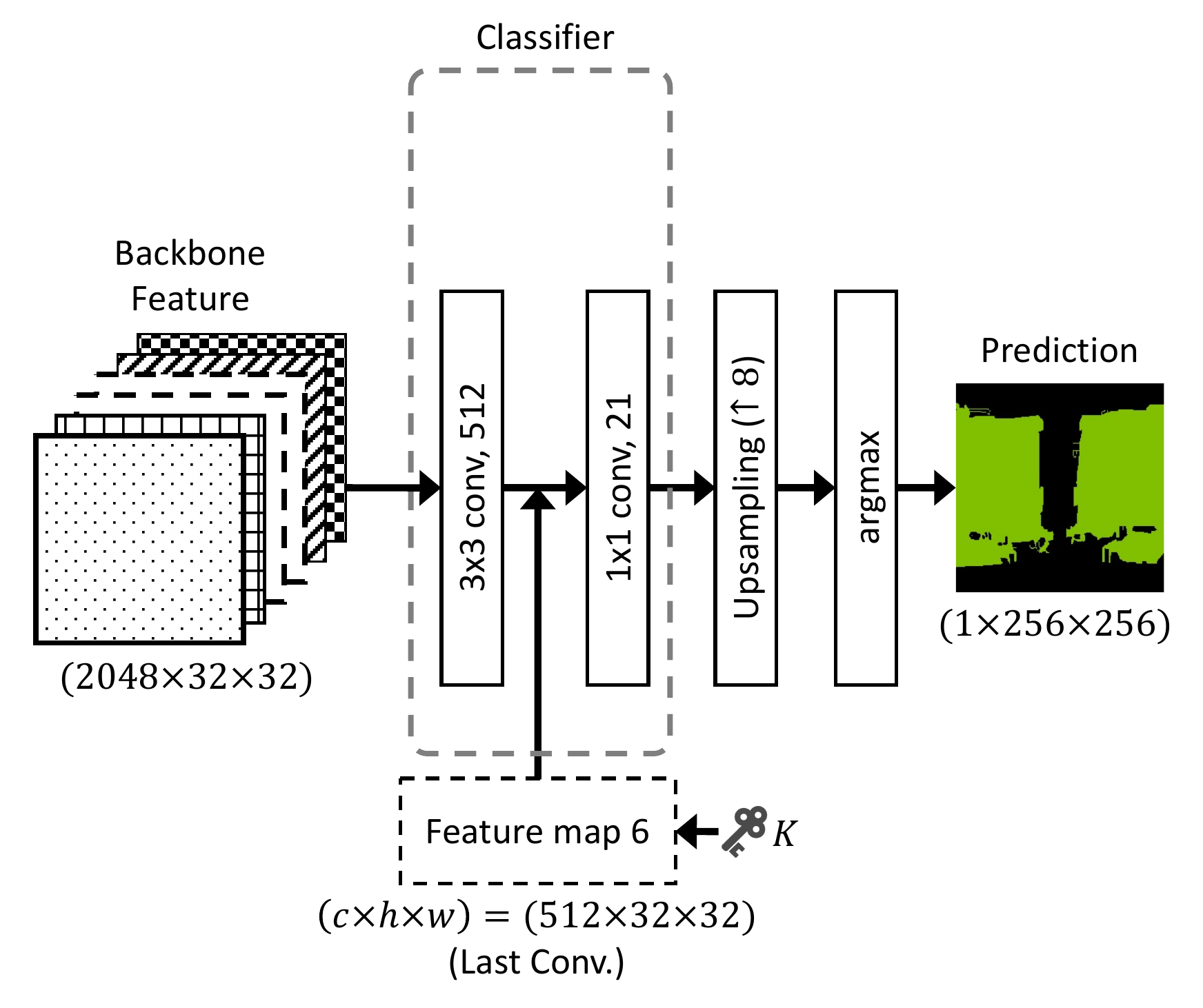}
        \caption{FCN classifier}
        \label{fig:fcn_head}
    \end{subfigure}
    \begin{subfigure}{9cm}
        \centering
        \includegraphics[height=4cm]{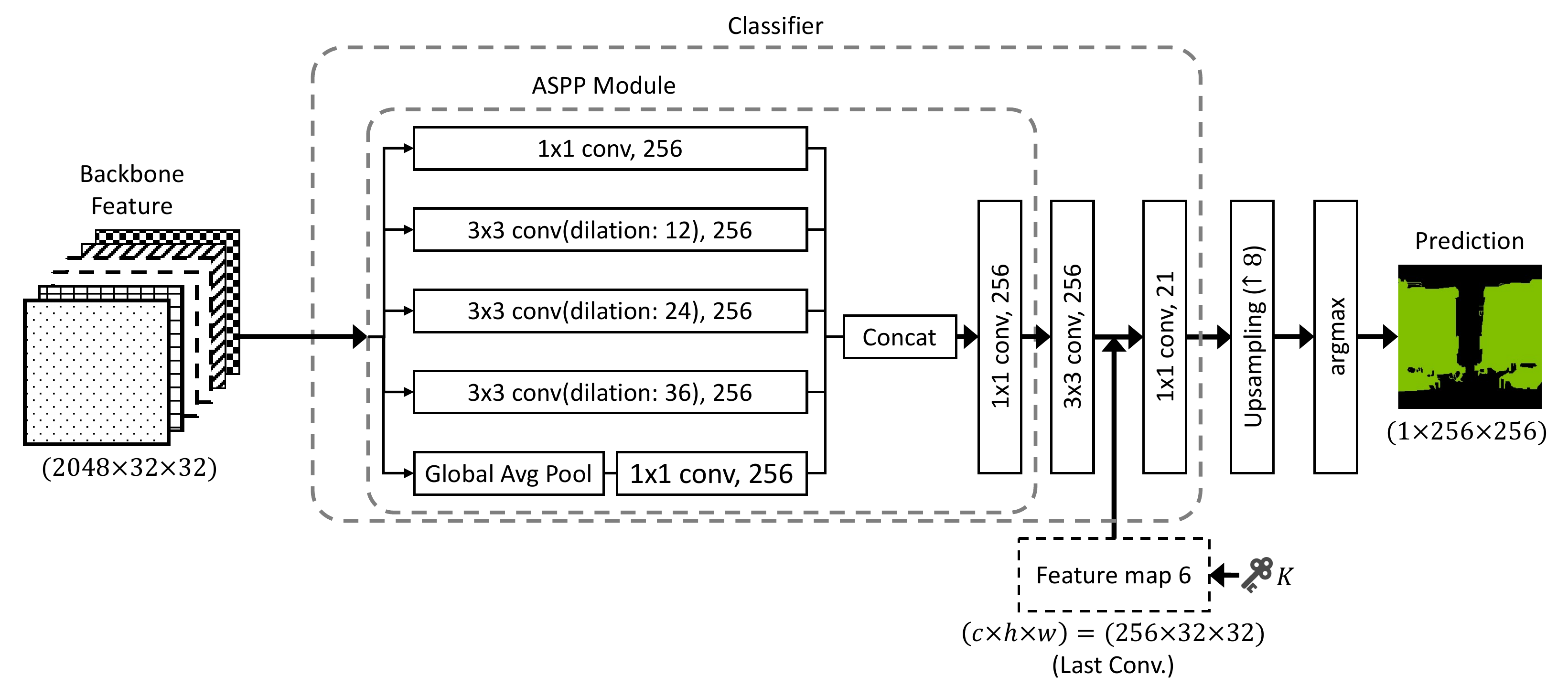}
        \caption{DeepLabv3 classifier}
        \label{fig:deeplab_head}
    \end{subfigure}
    \caption{Segmentation models using encrypted feature maps}
    \label{fig:model}
\end{figure*}

In the proposed method, one or more feature maps in a network are selected, and then the selected feature maps are encrypted with a secret key.
We illustrate semantic segmentation architectures in Fig.~\ref{fig:model} as an example, in which there are six feature maps (feature maps 1-6) where two classifiers correspond to a fully convolutional network (FCN) \cite{fcn} and a network using atrous convolution (DeepLabv3) \cite{deeplab}, respectively.
Both networks consist of one backbone and one classifier, and ResNet-50 is commonly used as the backbone.
Input images are fed to the backbone, and the classifier gets features from the backbone.
Finally, the classifier outputs a prediction result.

\subsection{Feature Map Encryption}
\label{subsec:enc_method}
\RA{A feature map is an intermediate output in a convolutional network.
Unlike weights which are learned  by using all input images in a model, a feature map is decided by using each input image. Therefore, } 
a selected feature map $x \in \mathbb{R}^{c \times h \times w}$ is transformed with key $K$ at each iteration for training a model, where $c$ is the number of channels, $h$ is the height, and $w$ is the width of the feature map.
To transform feature maps, we address two methods: pixel shuffling and channel permutation as follows.

\subsubsection{Block-Wise Pixel Shuffling}
A block-wise pixel shuffling method, referred to as pixel shuffling (SHF), was investigated as a method for encrypting input images in \cite{access3,adv-def}.
In this paper, SHF is extended for the access control of semantic segmentation models.

\begin{figure*}
    \centering
    \includegraphics[width=0.99\textwidth]{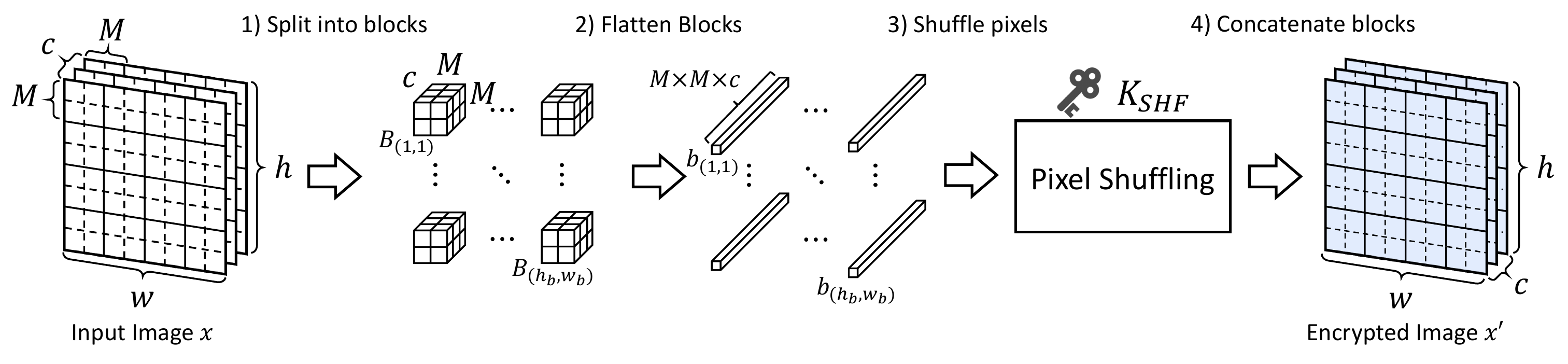}
    \caption{Block-wise pixel shuffling process for input image encryption}
    \label{fig:block_trans}
\end{figure*}

Below is the encryption procedure of the conventional SHF (see Fig.~\ref{fig:block_trans}), where $x$ is an input image.
\begin{enumerate}
    \item Divide $x$ into blocks with a size of $M \times M$ as
    \begin{equation}
    \{ B_{(1,1)}, \ldots, B_{(l,m)}, \ldots, B_{(h_b, w_b)} \},
    \end{equation}
    where $h_b \times w_b$ denotes the number of blocks, and each block has a shape of ($c,M,M$).
    \item Flatten each block $B_{(l,m)}$ as a vector
    \begin{equation}
    b_{(l,m)} = [{b_{(l,m)}}(1), \ldots, {b_{(l,m)}}(L)],
    \end{equation}
    where the length of the flattened vector is $L = c \times M \times M$.
    \item Shuffle pixels: First, generate secret key $K_{\mathrm{SHF}}$ as
    \begin{equation}
    K_{\mathrm{SHF}} = [\alpha_1, \ldots, \alpha_i, \ldots, \alpha_{i'}, \ldots, \alpha_L],
    \end{equation}
    where $\alpha_i \in \{1, \ldots, L\}$, and $\alpha_i \neq \alpha_{i'}$ if $i \neq i'$.
    Second, shuffle each vector $b_{(l,m)}$ with $K_{\mathrm{SHF}}$ such that
    \begin{equation}
    b'_{(l,m)}(i) = b_{(l,m)}(\alpha_i),
    \end{equation}
    and a shuffled vector is given by
    \begin{equation}
    b'_{(l,m)} = [{b'_{(l,m)}}(1), \ldots, {b'_{(l,m)}}(L)].
    \end{equation}
    All vectors are converted with the same key.
    \item Concatenate blocks: The shuffled vectors are integrated to obtain transformed input image $x'$ with a dimension of $(c, h, w)$.
\end{enumerate}

\subsubsection{Channel Permutation}
SHF is extended for application to semantic segmentation models in terms of two points: the use of feature maps and a block size of $M=1$.
The extended encryption is called channel permutation (CP).
Accordingly, CP is a pixel-wise transformation, where a feature map is permuted only along the channel dimension.

\begin{figure}
    \centering
    \includegraphics[width=0.7\columnwidth]{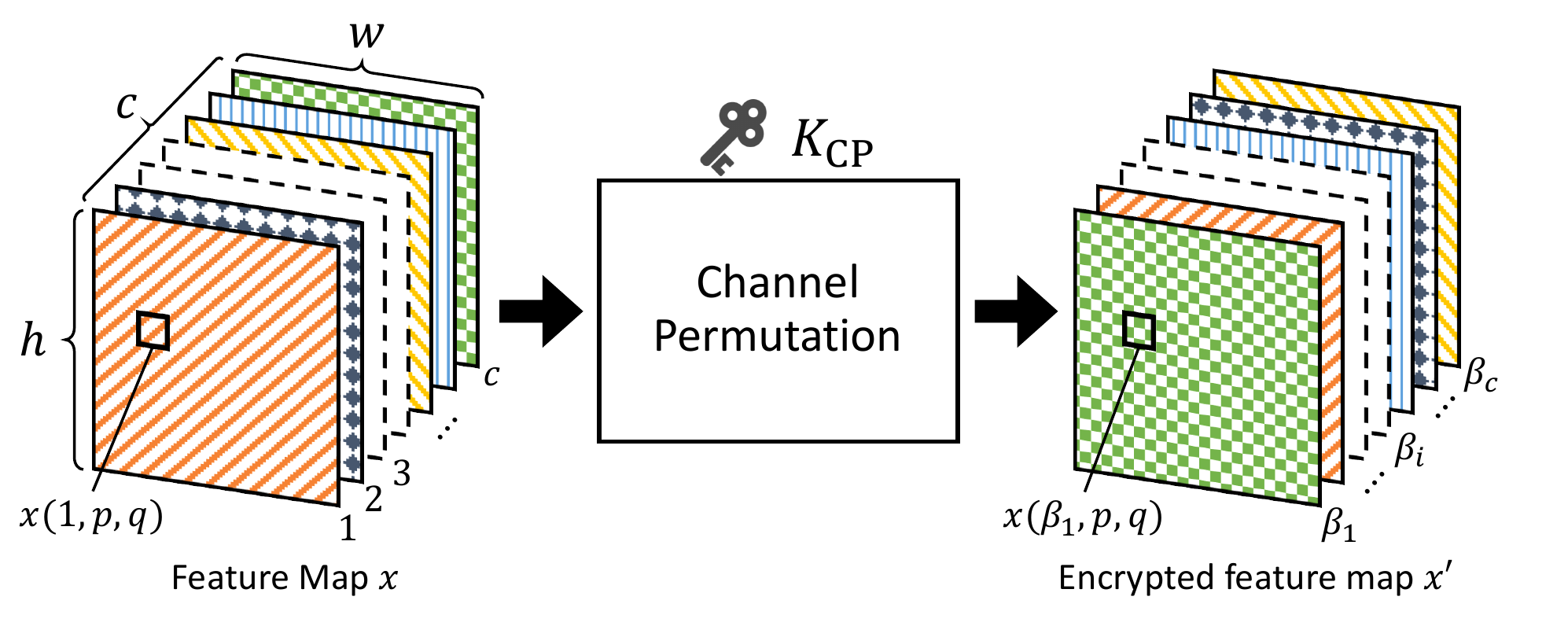}
    \caption{Channel permutation process}
    \label{fig:perm_trans}
\end{figure}

The following is the procedure of CP.
\begin{enumerate}
    \item Select a feature map $x$ to be encrypted.
    \item Generate secret key $K_{\mathrm{CP}}$ with a size of $c$ as
    \begin{equation}
    K_{\mathrm{CP}} = [\beta_1, \ldots, \beta_j, \ldots, \beta_{j'}, \ldots, \beta_c],
    \end{equation}
    where $\beta_j \in \{1, \ldots, c\}$, and $\beta_j \neq \beta_{j'}$ if $j \neq j'$.
    \item Replace all elements of $x$, $x(j, p, q)$, $p \in \{1,\ldots,h\}$, and $q \in \{1,\ldots,w\}$ as
    \begin{equation}
    x'(j,p,q) = x(k_j,p,q),
    \end{equation}
    and permuted feature map $x' \in \mathbb{R}^{c \times h \times w}$ is obtained.
\end{enumerate}
As shown in Fig.~\ref{fig:perm_trans}, CP is a spatially-invariant transformation, so it can support a pixel-level resolution, which is important for semantic segmentation, even though SHF supports a block-level one.

\subsection{Difference between SHF and CP}
The differences between SHF and CP are summarized as below.
\begin{enumerate}[label=(\alph*)]
\item CP is a pixel-wise transformation.
\item CP is applied to a feature map.
\item The number of feature map channels is larger than that of input image channels.
\end{enumerate}
Difference (a) allows us to obtain results with a pixel-level resolution, but pixel-wise transformations are not robust against various attacks if the transformation is applied to input images as discussed in \cite{access3} because the number of input image channels $c$ is small (i.e., RGB images have $c=3$).
To improve on this, we propose encrypting feature maps that have a larger number of channels such as $c=2048$ as shown in Fig.~\ref{fig:model}.
In addition, the use of feature maps enables us to maintain a high accuracy as described later.

\subsection{Threat Models}
A threat model includes a set of assumptions such as an attacker's goals, knowledge, and capabilities.
Users without secret key $K$ are assumed to be the adversary.
Attackers may steal a model to achieve different goals for profit.
In this paper, we consider the attacker's goal is to be able to make use of a stolen model.
This paper considers brute-force, random key, and fine-tuning attacks as ciphertext-only attacks.
Therefore, the following possible attacks done with the intent of stealing a model are discussed to evaluate the robustness of the proposed model-protection method.
In experiments, the method will be demonstrated to be robust against attacks.

\subsubsection{Brute-Force Attack}
A simple attack to decrypt an encrypted input image or feature map is a brute-force attack.
This attack systematically checks all possible passwords until the correct one is found.
Therefore, the encryption method must have a large enough key space.
The key space of each method is summarized in Table~\ref{tab:key space}.
\begin{table}[t]
    \centering
	\caption{Key space of access control methods}
	\label{tab:key space}
    \begin{tabular}{lcc} \hline \hline
    Method & Key space & Remark \\ \hline
    SHF & $(c \times M \times M)!$ & $c=3$ (input image) \\
    CP & $c!$ & $c \gg 3$ (feature map) \\ \hline \hline
    \end{tabular}
\end{table}
The key space is decided by block size $M$ and channel size $c$.
For example, if SHF with $M=2$ is applied to an input image, the key space is $(3\times2\times2)! \approx 4.79 \times 10^8 < 2^{29}$.
In contrast, when a feature map with $c=256$ is encrypted by using CP, the key space is $256! \approx 2^{1684}$.
In general, the channel number of feature maps is much larger than that of input images, so CP has a large key space even when $M=1$ is selected, compared with SHF.
In addition, when using CP, the attacker has to know or estimate the location of the transformed feature map, which cannot be known from the model itself.

\subsubsection{Random Key Attack}
\label{subsubsec:random}
In reality, the random attack is hard to carry out for the proposed model protection because there are many layers in a conventional CNN architecture, and the location of the transformed feature map cannot be known from the model itself.
To be practical, the cost of an attack should always be lower than that of training a new model.
We will consider a worst-case scenario in which an attacker obtains additional information about the transformed feature map and the transformation process except for the secret key, in an experiment.

\subsubsection{Fine-Tuning Attack}
\label{subsubsec:fine}
Fine-tuning is a process that takes a trained model and then tunes the model to make it perform some purpose (e.g., to process a similar task).
An attacker may use fine-tuning as an attack to override model protection so that the attacker can utilize a protected model without a secret key.
This attack aims to disable the key by retraining a protected model with a small subset of a dataset.
We assume an attacker has the model weights and a small dataset $D'$ for this attack.

\section{Experimental Results}
\label{sec:experiment}
To verify the effectiveness of the proposed method, the method was evaluated in terms of access control and robustness against attacks.
All experiments were conducted with the PyTorch library \cite{pytorch} in Python.

\subsection{Setup}
\label{subsec:setup}
\subsubsection{Dataset}
Semantic segmentation models were trained by using a dataset released for the segmentation competition of Visual Object Classes Challenge 2012 (VOC2012) \cite{pascal}.
The dataset consists of a training set with 1464 pairs (i.e., images and corresponding ground truths) and a development set with 1449 pairs.
In addition, a test set is also available only on the evaluation server, but it was not used in the experiment due to some constraints.

The training set was divided into 1318 samples for training models and 146 samples for validating the loss of models during the training, and we selected the model that provided the lowest loss value after the training.
The performance of the trained models was evaluated by using the development set with 1449 pairs.

All input images and ground truths were resized to a size of $256 \times 256$ because block-wise transformation requires images with a fixed size.
In addition, standard data-augmentation methods, i.e., random resized crop and horizontal flip, were performed in training models.

\subsubsection{Networks}
We used a fully convolutional network (FCN) \cite{fcn} and a network with atrous convolution (DeepLabv3) \cite{deeplab} for semantic segmentation, as shown in Fig.~\ref{fig:model}.
In the experiments, a deep residual network with 50 layers (ResNet-50) \cite{resnet} was used as a backbone for both networks, where only the backbone was pre-trained on a dataset used in ImageNet Large Scale Visual Recognition Challenge 2012 (ILSVRC2012) \cite{imagenet}, and the pre-trained weights were provided on PyTorch.
All networks were trained for 30 epochs by using a stochastic gradient descent (SGD) optimizer, where an initial learning rate (lr) of 0.02, a weight decay of 0.0001, and a momentum of 0.9 were selected as the hyperparameters of the optimizer.
The learning rate was decayed in each iteration as
\begin{equation}
    lr = 0.02 \times \left( 1-\frac{n}{30 \times 42} \right)^{0.9},
\end{equation}
where $n$ is the current iteration number.
The batch size was 32, and the standard pixel-wise cross-entropy loss without weight rebalancing was used.

\subsection{Performance Evaluation}
In this experiment, the segmentation performance of the protected models was evaluated with the mean intersection-over-union (mean IoU), which is a common evaluation metric for semantic segmentation.
An IoU value is given for each class by
\begin{equation}
    \label{eq:iou}
    \mathrm{IoU} = \frac{TP}{TP+FP+FN}\quad,
\end{equation}
and the mean IoU is then calculated by averaging the IoU values of all classes.
$TP$, $FP$, and $FN$ mean true positive, false positive, and false negative values calculated from predicted segmentation maps and ground truth ones, respectively.
In addition, the metric ranges from zero to one, where a value of one means that the predicted segmentation maps are the same as those of the ground truths, and a value of zero indicates that they have no overlap.

\subsubsection{Model Trained with Encrypted Feature Map}
\label{subsubsec:feature_performance}
\begin{figure*}
    \centering
    \begin{tabular}{ccccc}
     & Input & Correct & No-enc & Groud Truth \\
    \parbox[t]{1mm}{\rotatebox[origin=c]{90}{Example 1}} &
    \begin{minipage}{2cm}
        \centering
        \includegraphics[width=2cm]{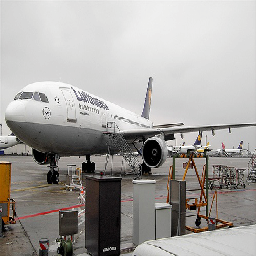}
    \end{minipage} &
    \begin{minipage}{2cm}
      \centering
      \includegraphics[width=2cm]{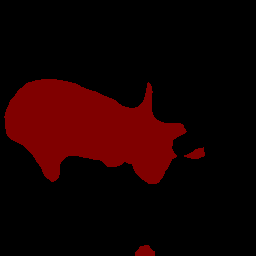}
    \end{minipage} &
    \begin{minipage}{2cm}
      \centering
      \includegraphics[width=2cm]{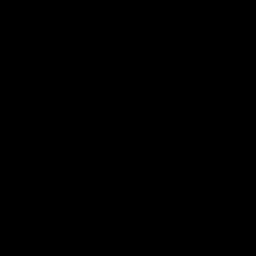}
    \end{minipage} &
    \begin{minipage}{2cm}
        \centering
        \includegraphics[width=2cm]{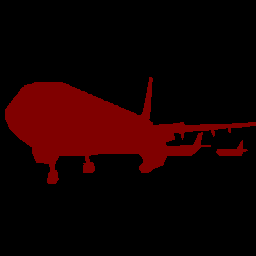}
    \end{minipage} \\
     & & 88.67 & 41.16 & \\
    \parbox[t]{1mm}{\rotatebox[origin=c]{90}{Example 2}} &
    \begin{minipage}{2cm}
        \centering
        \includegraphics[width=2cm]{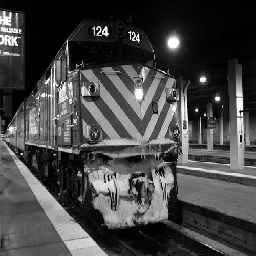}
    \end{minipage} &
    \begin{minipage}{2cm}
      \centering
      \includegraphics[width=2cm]{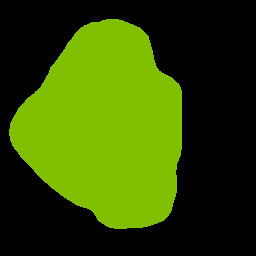}
    \end{minipage} &
    \begin{minipage}{2cm}
      \centering
      \includegraphics[width=2cm]{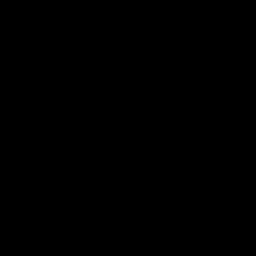}
    \end{minipage} &
    \begin{minipage}{2cm}
        \centering
        \includegraphics[width=2cm]{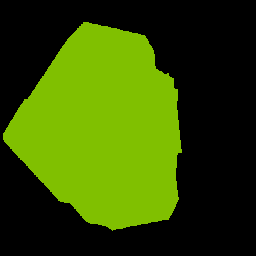}
    \end{minipage} \\
     & & 95.88 & 29.84 & \\
    \parbox[t]{1mm}{\rotatebox[origin=c]{90}{Example 3}} &
    \begin{minipage}{2cm}
        \centering
        \includegraphics[width=2cm]{./figure/pred_sample_image/F-CP/in_23.png}
    \end{minipage} &
    \begin{minipage}{2cm}
      \centering
      \includegraphics[width=2cm]{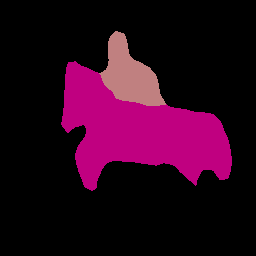}
    \end{minipage} &
    \begin{minipage}{2cm}
      \centering
      \includegraphics[width=2cm]{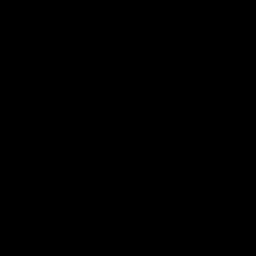}
    \end{minipage} &
    \begin{minipage}{2cm}
        \centering
        \includegraphics[width=2cm]{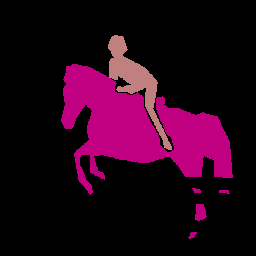}
    \end{minipage} \\
     & & 83.18 & 26.66 & \\
    \end{tabular}
    \caption{Example of prediction results for CP (DeepLabv3). Mean IoU values are given under predictions.}
    \label{fig:ex_sample_feature}
\end{figure*}

In this experiment, a channel permutation (CP) was applied to a selected feature map in a network for semantic segmentation.
Table~\ref{tab:feature_accuracy} shows the results under two classifiers: FCN and DeepLabv3, where one feature map was selected to be encrypted from six feature maps in each network (see Fig.~\ref{fig:model}).
In the table, the segmentation accuracy was calculated by using 1449 pairs under two conditions: Correct and No-enc, where ``Correct'' means the use of test images encrypted with correct key $K$, and ``No-enc'' indicates the use of plain test images.
An example of the results with DeepLabv3 is also shown in Fig.~\ref{fig:ex_sample_feature}.
\begin{table}[t]
\centering
\caption{\RA{Segmentation accuracy (mean IoU) of proposed method (CP). Best accuracies are shown in bold.}}
\label{tab:feature_accuracy}
\begin{tabular}{ll|cc|cc} \hline \hline
\multicolumn{2}{l|}{Network} & \multicolumn{2}{c|}{FCN} & \multicolumn{2}{c}{DeepLabv3} \\ \hline
\multicolumn{2}{l|}{Key condition} & Correct            & No-enc           & Correct         & No-enc \\ \hline
\multirow{6}{*}{\makecell{Selected \\ feature \\ map}} & 1                                        & 46.35              & 14.71              & 54.79           & 15.82 \\
 & 2                                        & 43.86              & 29.43              & 51.38           & 37.82 \\
 & 3                                        & 34.18              & 7.76               & 38.72           & 10.06 \\
 & 4                                        & 50.79              & 3.79               & 55.33           & 3.93 \\
 & 5                                        & 57.19              & 3.80               & 64.75           & 3.57 \\
 & 6                                        & \textbf{58.52}              & \textbf{3.49}               & \textbf{65.15}           & \textbf{3.49} \\ \hline
\multicolumn{2}{l|}{Baseline} & \multicolumn{2}{c|}{58.89 (non-protected)} & \multicolumn{2}{c}{65.77 (non-protected)} \\ \hline \hline
\end{tabular}
\end{table}

From the table, CP was confirmed to achieve almost the same accuracy as that of the baselines under the use of the correct key when feature map 5 or 6 was selected.
In contrast, CP provided a low accuracy to unauthorized users without the key (No-enc).
\RA{
Note that the segmentation performance slightly varies in general due to the initial weights of a model and the key. We carried out the experiment 10 times with different initial weights and keys under each condition. Average results were presented in Table~\ref{tab:feature_accuracy}. From the experiments, we confirmed that the proposed access control method with a selected feature map encryption can achieve almost the same performance as the baseline (non-protected) model.
}

From Fig.~\ref{fig:ex_sample_feature}, the prediction results were confirmed to be similar to the corresponding ground truths under the use of the correct key.
In contrast, the results estimated from plain images had only a background label.
Accordingly, CP with encrypted feature maps was effective in the access control of semantic segmentation models.

\subsubsection{Selection of Feature Maps}

As shown in Table~\ref{tab:feature_accuracy}, the performance of the models trained with encrypted feature maps depended on the selection of feature maps.
In the experiment, when the encryption was applied to a feature map at positions 2 to 4, the segmentation accuracy was lower than that of models 5 and 6.
The difference in segmentation accuracy among the selected feature maps was caused by a residual connection in the ResNet-50 backbone in Fig.~\ref{fig:model}.
From Fig.~\ref{fig:model}, feature maps 2-4 had residual connections on both the front and back of each feature map.
In contrast, in feature maps 1, 5, and 6, the influence of CP can be easily canceled out by a convolutional layer because there is no residual connection either in front or behind.
Accordingly, feature map 6 is recommended as an encrypted feature map.

Although the access control performance of the models trained with encrypted feature maps depend upon the selection of feature maps, the selection of feature maps is independent of the type of datasets. Accordingly, we can experimentally select a feature map to be encrypted under the use of a dataset. \RA{In principle, one or more feature maps can be encrypted in the proposed access control method. However, when unsuitable feature maps are encrypted, it degrades the performance of models as shown in Table 2. Our experiments confirmed that encrypting only one feature map has already provided a good access control performance, and encrypting two or more feature maps does not have any significant advantage. Therefore, only one feature map was
encrypted in experiments.}

\subsubsection{Model Trained with Encrypted Input Images}
Input images were encrypted in accordance with SHF under various block sizes (i.e., $M \in \{ 1,2,4,8,16,32 \}$) for comparison with the proposed method (CP).
SHF was already demonstrated to achieve a high access control performance in image classification tasks in \cite{access3}, but it has never been applied to semantic segmentation ones.

Table~\ref{tab:input_accuracy} shows the segmentation accuracy of the protected models calculated from 1449 pairs.
From the results, even when correct key $K$ was used, the segmentation accuracy decreased significantly as block size $M$ increased in both networks.
In contrast, when the block size was small, the protected model achieved a segmentation accuracy close to the baseline.
However, the accuracy without the encryption (i.e., No-enc) was almost the same as that of ``Correct,'' so the access control was weak under the use of a small block size.
\begin{table}[t]
\centering
\caption{Segmentation accuracy (mean IoU) of conventional method with encrypted input images (SHF)}
\label{tab:input_accuracy}
\begin{tabular}{ll|cc|cc} \hline \hline
\multicolumn{2}{l|}{Network} & \multicolumn{2}{c|}{FCN}                   & \multicolumn{2}{c}{DeepLabv3} \\ \hline
\multicolumn{2}{l|}{Key Condition} & Correct             & No-enc            & Correct             & No-enc            \\ \hline
\multirow{6}{*}{\makecell{Block \\ size \\ $M$}} & 1 & 56.55 & 56.15 & 64.76 & 62.88 \\
 & 2                           & 51.54               & 47.58               & 59.74               & 56.67               \\
 & 4                           & 48.37               & 46.72               & 50.82               & 51.96               \\
 & 8                           & 34.25               & 34.68               & 37.70               & 35.95               \\
 & 16                          & 18.05               & 13.42               & 20.91               & 15.83               \\
 & 32                          & 7.68                & 5.21                & 11.14               & 5.58                \\ \hline
\multicolumn{2}{l|}{Baseline} & \multicolumn{2}{c|}{58.89 (non-protected)} & \multicolumn{2}{c}{65.77 (non-protected)} \\ \hline \hline
\end{tabular}
\end{table}

\begin{figure*}
    \centering
    \begin{tabular}{c|cccccc}
    Block size & \multicolumn{2}{c}{$M=1$} & \multicolumn{2}{c}{$M=2$} & \multicolumn{2}{c}{$M=8$} \\ \hline
    Key condition & Correct & No-enc & Correct & No-enc & Correct & No-enc \\ \hline
     & & & & & & \\[-8pt]
    Input &
    \begin{minipage}{2cm}
      \centering
      \includegraphics[width=2cm]{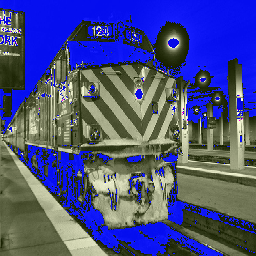}
    \end{minipage} &
    \begin{minipage}{2cm}
      \centering
      \includegraphics[width=2cm]{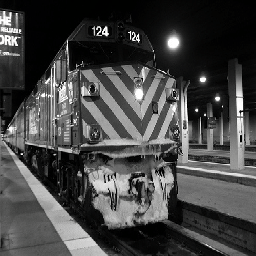}
    \end{minipage} &
    \begin{minipage}{2cm}
      \centering
      \includegraphics[width=2cm]{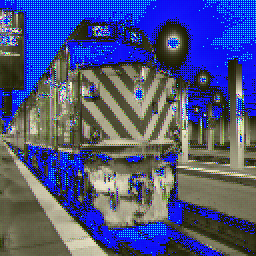}
    \end{minipage} &
    \begin{minipage}{2cm}
      \centering
      \includegraphics[width=2cm]{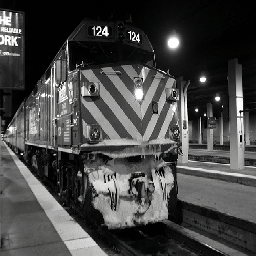}
    \end{minipage} &
    \begin{minipage}{2cm}
      \centering
      \includegraphics[width=2cm]{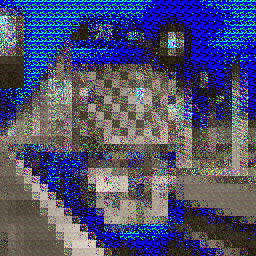}
    \end{minipage} &
    \begin{minipage}{2cm}
      \centering
      \includegraphics[width=2cm]{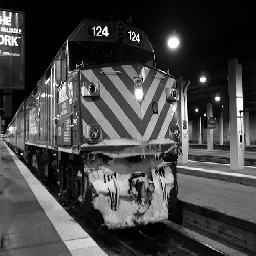}
    \end{minipage} \\
     & & & & & & \\[-8pt]
    Prediction &
    \begin{minipage}{2cm}
      \centering
      \includegraphics[width=2cm]{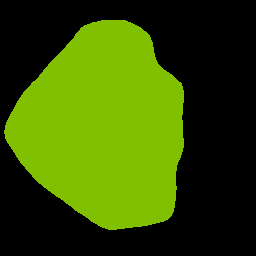}
    \end{minipage} &
    \begin{minipage}{2cm}
      \centering
      \includegraphics[width=2cm]{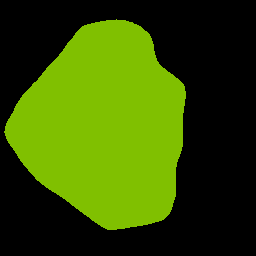}
    \end{minipage} &
    \begin{minipage}{2cm}
      \centering
      \includegraphics[width=2cm]{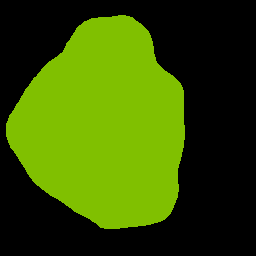}
    \end{minipage} &
    \begin{minipage}{2cm}
      \centering
      \includegraphics[width=2cm]{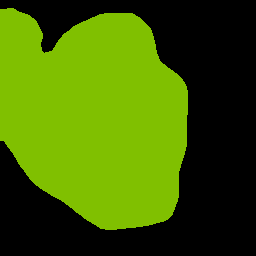}
    \end{minipage} &
    \begin{minipage}{2cm}
      \centering
      \includegraphics[width=2cm]{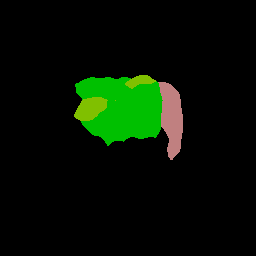}
    \end{minipage} &
    \begin{minipage}{2cm}
      \centering
      \includegraphics[width=2cm]{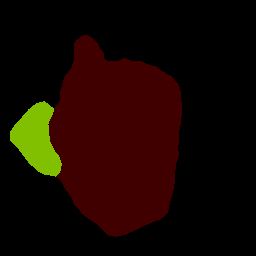}
    \end{minipage} \\
     & 97.33 & 96.80 & 96.18 & 83.89 & 17.21 & 32.51 \\
    Ground Truth &
    \multicolumn{6}{c}{
        \begin{minipage}{2cm}
          \centering
          \includegraphics[width=2cm]{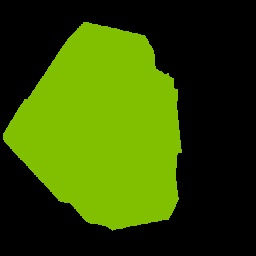}
        \end{minipage}
    }
    \end{tabular}
    \caption{Example of prediction results for SHF (DeepLabv3). Mean IoU values are given under predictions.}
    \label{fig:ex_sample_image}
\end{figure*}

From Fig.~\ref{fig:ex_sample_image}, we also confirmed that the prediction results for Correct were similar to those for No-enc when a small block size was used.
In addition, the prediction result for $M=8$ was significantly degraded compared with the ground truth.
Therefore, applying SHF to input images is not suitable for the access control of semantic segmentation models, even though it is suitable for image classification tasks.

\subsection{Robustness against Random Key Attack}
In this experiment, CP was evaluated in terms of robustness against the random key attack described in Section \ref{subsubsec:random}, where models were protected by encrypting feature map 6.
An evaluation was carried out on robustness with 100 incorrect keys that were randomly generated.

\begin{figure*}[t]
    \centering
    \begin{subfigure}{0.49\textwidth}
        \centering
        \includegraphics[width=0.95\columnwidth]{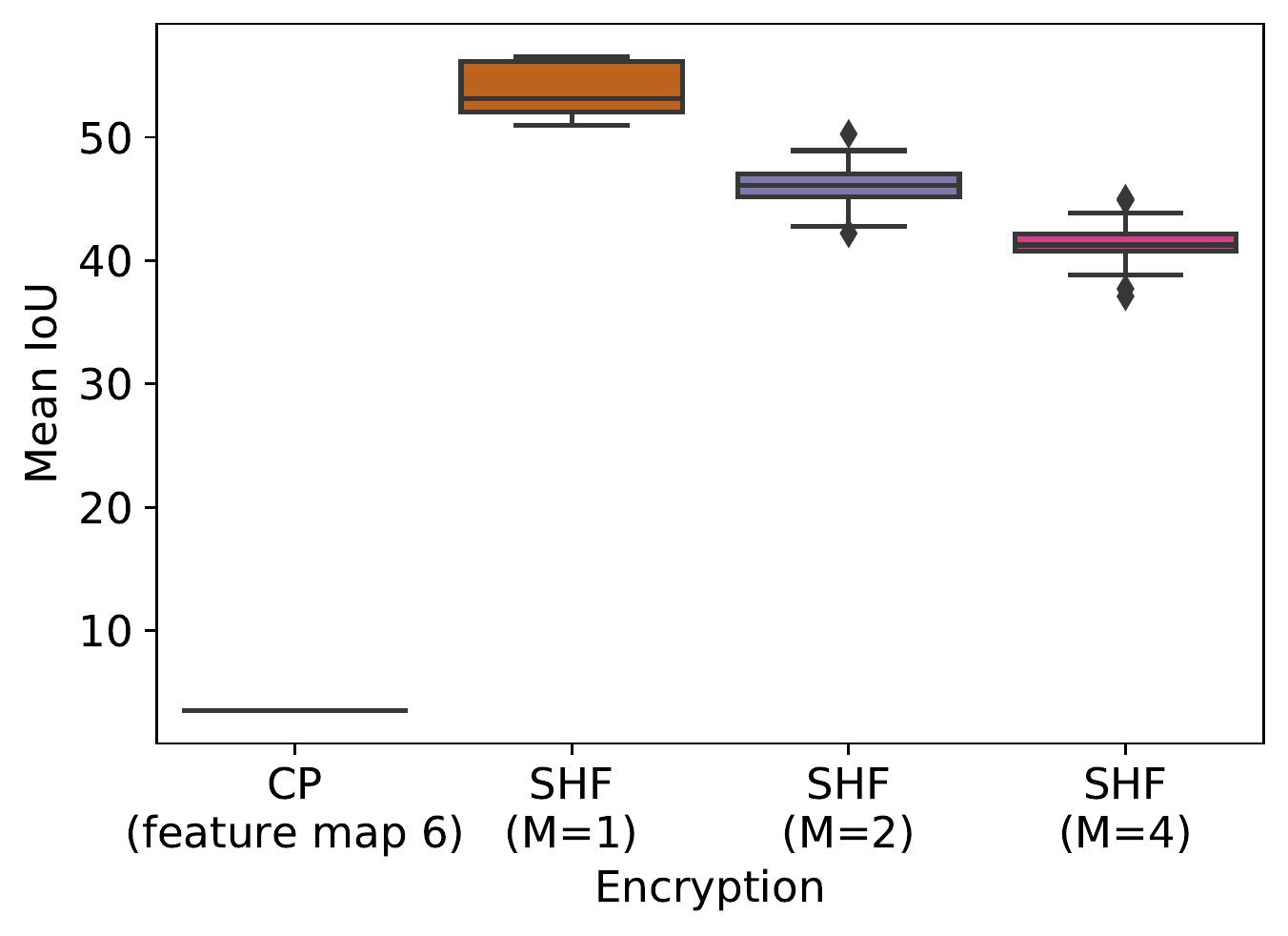}
        \caption{FCN}
        \label{fig:fcn_random}
    \end{subfigure}
    \begin{subfigure}{0.49\textwidth}
        \centering
        \includegraphics[width=0.95\columnwidth]{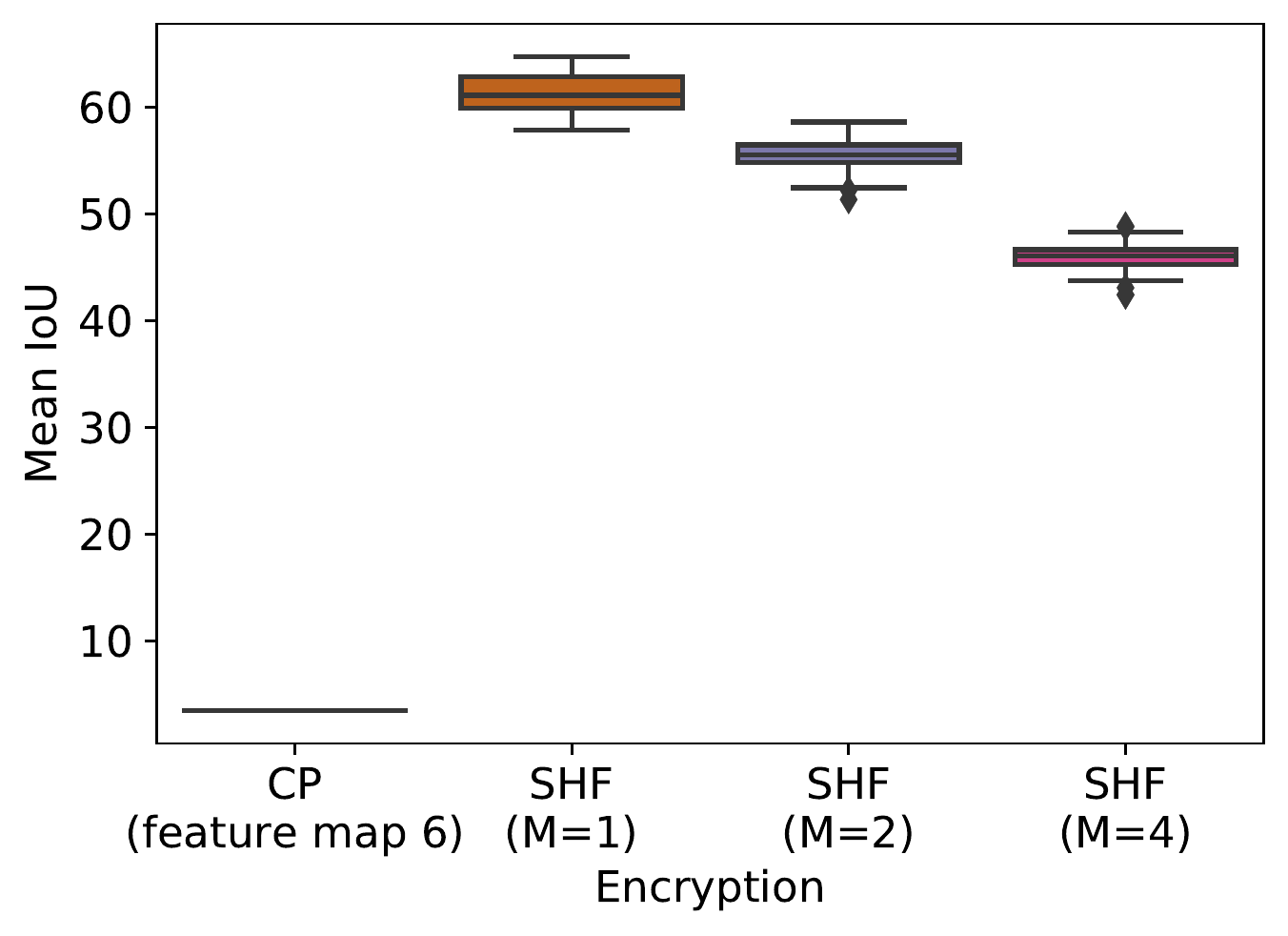}
        \caption{DeepLabv3}
        \label{fig:deep_random}
    \end{subfigure}
    \caption{Mean IoU values of protected models with 100 incorrect keys. Boxes span from first to third quartile, referred to as $Q_1$ and $Q_3$, and whiskers show maximum and minimum values in range of [$Q_1 - 1.5(Q_3 - Q_1), Q_3 + 1.5(Q_3 - Q_1)$]. Band inside box indicates median. Outliers are indicated as dots.}
    \label{fig:random}
\end{figure*}

Figure \ref{fig:random} shows the segmentation performance of the protected models under the use of the incorrect keys on the development set of VOC2012.
From the results of using CP, the mean IoU values were significantly low for both models, which means that the models were robust enough against this attack.
However, the mean IoU values of using SHF increased as the block size decreased.
Therefore, the proposed method (CP) outperformed the conventional method (SHF) in terms of robustness against the random key attack.

\subsection{Robustness against Fine-Tuning Attack}
We ran an experiment with different sizes for an attacker's small dataset (i.e., $|D'| \in \{5\%, 10\% \}$ of the training data).
Models protected by encrypting feature map 6 were retrained by using $D'$ to disable the key.
Table~\ref{tab:fine_accuracy} shows the results of the fine-tuning attack for both networks.

Although the accuracy of the fine-tuned models was higher when the size of $D'$ was larger, it was still lower than the accuracy of the original protected models (i.e., ``Protected'' in Table~\ref{tab:fine_accuracy}).
Therefore, the attacker was not able to use the models to full capacity even when preparing a small dataset.
\begin{table}[t]
\centering
\caption{Segmentation accuracy (mean IoU) of fine-tuned models}
\label{tab:fine_accuracy}
\begin{tabular}{lll|cc} \hline \hline
\multicolumn{3}{l|}{Network} & FCN & DeepLabv3 \\ \hline
\multirow{3}{*}{\makecell[l]{Fine-tuned \\ (test without key)}} & \multirow{3}{*}{$D'$} & 0\%  & 3.49 & 3.49\\
 & & 5\% & 27.02 & 29.78  \\
 & & 10\% & 41.70 & 46.19 \\ \hline
\multicolumn{3}{l|}{\makecell[l]{No fine-tuned \\ (test with key)}} & 59.24 & 65.43 \\ \hline \hline
\end{tabular}
\end{table}


\section{Conclusion and Future Work}
\label{sec:conclusion}
In this paper, we proposed an access control method for semantic segmentation models for the first time.
The method is carried out by encrypting selected feature maps with a secret key called channel permutation (CP), while input images are encrypted by using a block-wise encryption method in conventional methods.
The use of CP allows us not only to obtain a pixel-level accuracy that is required for semantic segmentation but also to maintain a wide key space even when a pixel-wise permutation is used.
As a result, the proposed method can maintain both a high accuracy and robustness against attacks.
In experiments, the conventional method with encrypted input images was not effective in the access control of semantic segmentation models, and the effectiveness of the proposed method was demonstrated in terms of segmentation accuracy.

\RB{As for future work, we shall generalize the proposed method to other models such as object detection models and generative models.} \RC{In addition, if the key is compromised, the proposed method in its current form needs to repeat the whole training to update the key in the same way that existing key-based access control methods such as the use of encrypted input images for image classification need to repeat the training. To overcome this limitation, we shall explore possible options in our future work.} \RB{We shall also identify other potential threats to the access control of the models.}


\bibliographystyle{IEEEtran}
\bibliography{refs}

\begin{thebibliography}{10}
\providecommand{\url}[1]{#1}
\csname url@samestyle\endcsname
\providecommand{\newblock}{\relax}
\providecommand{\bibinfo}[2]{#2}
\providecommand{\BIBentrySTDinterwordspacing}{\spaceskip=0pt\relax}
\providecommand{\BIBentryALTinterwordstretchfactor}{4}
\providecommand{\BIBentryALTinterwordspacing}{\spaceskip=\fontdimen2\font plus
\BIBentryALTinterwordstretchfactor\fontdimen3\font minus
  \fontdimen4\font\relax}
\providecommand{\BIBforeignlanguage}[2]{{%
\expandafter\ifx\csname l@#1\endcsname\relax
\typeout{** WARNING: IEEEtran.bst: No hyphenation pattern has been}%
\typeout{** loaded for the language `#1'. Using the pattern for}%
\typeout{** the default language instead.}%
\else
\language=\csname l@#1\endcsname
\fi
#2}}
\providecommand{\BIBdecl}{\relax}
\BIBdecl

\bibitem{dl1}
Y.~LeCun, Y.~Bengio, and G.~Hinton, ``Deep learning,'' \emph{Nature}, vol. 521,
  no. 7553, pp. 436--444, 5 2015.

\bibitem{application0}
X.~Liu, Z.~Deng, and Y.~Yang, ``Recent progress in semantic image
  segmentation,'' \emph{Artif. Intell. Rev.}, vol.~52, no.~2, pp. 1089--1106, 8
  2019.

\bibitem{kiya2022overview}
H.~Kiya, M.~AprilPyone, Y.~Kinoshita, S.~Imaizumi, and S.~Shiota, ``An overview
  of compressible and learnable image transformation with secret key and its
  applications,'' \emph{APSIPA Transactions on Signal and Information
  Processing}, vol.~11, no.~1, p. e11, 2022.

\bibitem{watermark}
M.~Swanson, M.~Kobayashi, and A.~Tewfik, ``Multimedia data-embedding and
  watermarking technologies,'' \emph{Proceedings of the IEEE}, vol.~86, no.~6,
  pp. 1064--1087, 1998.

\bibitem{embed0}
M.~Xue, J.~Wang, and W.~Liu, ``{DNN} intellectual property protection:
  Taxonomy, attacks and evaluations (invited paper),'' in \emph{Proceedings of
  the 2021 on Great Lakes Symposium on VLSI}, 6 2021, pp. 455--460.

\bibitem{embed1}
Y.~Uchida, Y.~Nagai, S.~Sakazawa, and S.~Satoh, ``Embedding watermarks into
  deep neural networks,'' in \emph{Proceedings of the 2017 ACM on International
  Conference on Multimedia Retrieval}, 2017, pp. 269--277.

\bibitem{embed3}
B.~Darvish~Rouhani, H.~Chen, and F.~Koushanfar, ``Deepsigns: An end-to-end
  watermarking framework for ownership protection of deep neural networks,'' in
  \emph{Proceedings of the Twenty-Fourth International Conference on ASPLOS},
  2019, pp. 485--497.

\bibitem{embed7}
H.~Chen, B.~D. Rouhani, C.~Fu, J.~Zhao, and F.~Koushanfar, ``Deepmarks: A
  secure fingerprinting framework for digital rights management of deep
  learning models,'' in \emph{Proceedings of the 2019 on International
  Conference on Multimedia Retrieval}.\hskip 1em plus 0.5em minus 0.4em\relax
  New York, NY, USA: Association for Computing Machinery, 2019, pp. 105--113.

\bibitem{embed5}
L.~Fan, K.~W. Ng, and C.~S. Chan, ``Rethinking deep neural network ownership
  verification: Embedding passports to defeat ambiguity attacks,'' in
  \emph{Advances in Neural Information Processing Systems}, vol.~32, 2019, pp.
  4716--4725.

\bibitem{embed4}
E.~Le~Merrer, P.~P{\'e}rez, and G.~Tr{\'e}dan, ``Adversarial frontier stitching
  for remote neural network watermarking,'' \emph{Neural Computing and
  Applications}, vol.~32, no.~13, pp. 9233--9244, 7 2020.

\bibitem{embed6}
S.~Sakazawa, E.~Myodo, K.~Tasaka, and H.~Yanagihara, ``Visual decoding of
  hidden watermark in trained deep neural network,'' in \emph{2019 IEEE
  Conference on Multimedia Information Processing and Retrieval}, 2019, pp.
  371--374.

\bibitem{embed9}
A.~MaungMaung and H.~Kiya, ``Piracy-resistant {DNN} watermarking by block-wise
  image transformation with secret key,'' in \emph{Proceedings of the 2021
  {ACM} Workshop on Information Hiding and Multimedia Security}.\hskip 1em plus
  0.5em minus 0.4em\relax Association for Computing Machinery, 2021, p.
  159–164.

\bibitem{embed2}
J.~Zhang, Z.~Gu, J.~Jang, H.~Wu, M.~P. Stoecklin, H.~Huang, and I.~Molloy,
  ``Protecting intellectual property of deep neural networks with
  watermarking,'' in \emph{Proceedings of the 2018 on Asia Conference on
  Computer and Communications Security}, 2018, pp. 159--172.

\bibitem{embed8}
Y.~Adi, C.~Baum, M.~Cisse, B.~Pinkas, and J.~Keshet, ``Turning your weakness
  into a strength: Watermarking deep neural networks by backdooring,'' in
  \emph{27th {USENIX} Security Symposium ({USENIX} Security 18)}, 8 2018, pp.
  1615--1631.

\bibitem{inv-attack}
M.~Fredrikson, S.~Jha, and T.~Ristenpart, ``Model inversion attacks that
  exploit confidence information and basic countermeasures,'' in
  \emph{Proceedings of the 22nd ACM SIGSAC Conference on Computer and
  Communications Security}, 2015, pp. 1322--1333.

\bibitem{adv1}
C.~Szegedy, W.~Zaremba, I.~Sutskever, J.~Bruna, D.~Erhan, I.~J. Goodfellow, and
  R.~Fergus, ``Intriguing properties of neural networks,'' in \emph{2nd
  International Conference on Learning Representations}, 2014.

\bibitem{adv2}
I.~J. Goodfellow, J.~Shlens, and C.~Szegedy, ``Explaining and harnessing
  adversarial examples,'' in \emph{3rd International Conference on Learning
  Representations}, 2015.

\bibitem{adv3}
N.~Papernot, P.~McDaniel, I.~Goodfellow, S.~Jha, Z.~B. Celik, and A.~Swami,
  ``Practical black-box attacks against machine learning,'' in
  \emph{Proceedings of the 2017 ACM on Asia Conference on Computer and
  Communications Security}, 2017, pp. 506--519.

\bibitem{access1}
M.~Chen and M.~Wu, ``Protect your deep neural networks from piracy,'' in
  \emph{2018 IEEE International Workshop on Information Forensics and
  Security}, 2018, pp. 1--7.

\bibitem{access3}
M.~AprilPyone and H.~Kiya, ``A protection method of trained {CNN} model with a
  secret key from unauthorized access,'' \emph{APSIPA Transactions on Signal
  and Information Processing}, vol.~10, p. e10, 2021.

\bibitem{adv-def}
------, ``Block-wise image transformation with secret key for adversarially
  robust defense,'' \emph{IEEE Transactions on Information Forensics and
  Security}, vol.~16, pp. 2709--2723, 2021.

\bibitem{pp1}
M.~Tanaka, ``Learnable image encryption,'' in \emph{2018 IEEE International
  Conference on Consumer Electronics-Taiwan (ICCE-TW)}, 2018, pp. 1--2.

\bibitem{pp2}
K.~Madono, M.~Tanaka, M.~Onishi, and T.~Ogawa, ``Block-wise scrambled image
  recognition using adaptation network,'' in \emph{Artificial Intelligence of
  Things (AIoT), Workshop on AAAI conference Artificial Intellignece,
  (AAAI-WS)}, 2020.

\bibitem{pp3}
W.~Sirichotedumrong and H.~Kiya, ``A gan-based image transformation scheme for
  privacy-preserving deep neural networks,'' in \emph{28th European Signal
  Processing Conference, EUSIPCO}, 2020, pp. 745--749.

\bibitem{pp5}
W.~Sirichotedumrong, Y.~Kinoshita, and H.~Kiya, ``Pixel-based image encryption
  without key management for privacy-preserving deep neural networks,''
  \emph{IEEE Access}, vol.~7, pp. 177\,844--177\,855, 2019.

\bibitem{pp6}
W.~Sirichotedumrong, T.~Maekawa, Y.~Kinoshita, and H.~Kiya,
  ``Privacy-preserving deep neural networks with pixel-based image encryption
  considering data augmentation in the encrypted domain,'' in \emph{2019 IEEE
  International Conference on Image Processing}, 2019, pp. 674--678.

\bibitem{etc1}
T.~Chuman, W.~Sirichotedumrong, and H.~Kiya, ``Encryption-then-compression
  systems using grayscale-based image encryption for jpeg images,'' \emph{IEEE
  Transactions on Information Forensics and Security}, vol.~14, no.~6, pp.
  1515--1525, 2019.

\bibitem{etc2}
W.~Sirichotedumrong and H.~Kiya, ``Grayscale-based block scrambling image
  encryption using ycbcr color space for encryption-then-compression systems,''
  \emph{APSIPA Transactions on Signal and Information Processing}, vol.~8,
  p.~e7, 2019.

\bibitem{etc4}
T.~Chuman, K.~Kurihara, and H.~Kiya, ``On the security of block
  scrambling-based {EtC} systems against extended jigsaw puzzle solver
  attacks,'' \emph{IEICE Transactions on Information and Systems}, vol. E101.D,
  no.~1, pp. 37--44, 2018.

\bibitem{fcn}
J.~Long, E.~Shelhamer, and T.~Darrell, ``Fully convolutional networks for
  semantic segmentation,'' in \emph{2015 IEEE Conference on Computer Vision and
  Pattern Recognition}, 2015, pp. 3431--3440.

\bibitem{deeplab}
\BIBentryALTinterwordspacing
L.-C. Chen, G.~Papandreou, F.~Schroff, and H.~Adam, ``Rethinking atrous
  convolution for semantic image segmentation,'' \emph{arXiv:1706.05587}, 2017.
  [Online]. Available: \url{https://arxiv.org/abs/1706.05587}
\BIBentrySTDinterwordspacing

\bibitem{pytorch}
A.~Paszke, S.~Gross, F.~Massa, A.~Lerer, J.~Bradbury, G.~Chanan, T.~Killeen,
  Z.~Lin, N.~Gimelshein, L.~Antiga, A.~Desmaison, A.~Kopf, E.~Yang, Z.~DeVito,
  M.~Raison, A.~Tejani, S.~Chilamkurthy, B.~Steiner, L.~Fang, J.~Bai, and
  S.~Chintala, ``Pytorch: An imperative style, high-performance deep learning
  library,'' in \emph{Advances in Neural Information Processing Systems 32},
  2019, pp. 8024--8035.

\bibitem{pascal}
M.~Everingham, S.~M.~A. Eslami, L.~Van~Gool, C.~K.~I. Williams, J.~Winn, and
  A.~Zisserman, ``The pascal visual object classes challenge: A
  retrospective,'' \emph{International Journal of Computer Vision}, vol. 111,
  no.~1, pp. 98--136, 1 2015.

\bibitem{resnet}
K.~He, X.~Zhang, S.~Ren, and J.~Sun, ``Deep residual learning for image
  recognition,'' in \emph{2016 IEEE Conference on Computer Vision and Pattern
  Recognition}, 2016, pp. 770--778.

\bibitem{imagenet}
O.~Russakovsky, J.~Deng, H.~Su, J.~Krause, S.~Satheesh, S.~Ma, Z.~Huang,
  A.~Karpathy, A.~Khosla, M.~Bernstein, A.~C. Berg, and L.~Fei-Fei, ``{ImageNet
  Large Scale Visual Recognition Challenge},'' \emph{International Journal of
  Computer Vision}, vol. 115, no.~3, pp. 211--252, 2015.

\end{thebibliography}
\end{document}